\pdfoutput=1

\documentclass[11pt]{article}

\usepackage[]{ACL2023}

\usepackage{times}
\usepackage{latexsym}

\usepackage[T1]{fontenc}

\usepackage[utf8]{inputenc}

\usepackage{microtype}

\usepackage{inconsolata}

\usepackage{graphicx}
\graphicspath{ {./figures/} }

\usepackage{multirow}

\usepackage[normalem]{ulem}
\useunder{\uline}{\ul}{}

\title{A Better Choice: Entire-space Datasets for Aspect Sentiment Triplet Extraction}

\author{Yuncong Li \\
	Tencent Inc \\
	Shenzhen, China \\
	\texttt{yuncongli@tencent.com} \\\And
	Fang Wang \\
	Shenzhen University \\
	Shenzhen, China \\
	\texttt{2160230414@email.szu.edu.cn} \\\And
	Sheng-hua Zhong \renewcommand{\thefootnote}{\fnsymbol{footnote}}\footnotemark[1] \\
	Shenzhen University \\
	Shenzhen, China \\
	\texttt{csshzhong@szu.edu.cn} \\
}

\begin{document}
\maketitle

\renewcommand{\thefootnote}{\fnsymbol{footnote}}
\footnotetext[1]{Corresponding author}
\renewcommand{\thefootnote}{\arabic{footnote}}

\begin{abstract}
	Aspect sentiment triplet extraction (ASTE) aims to extract aspect term, sentiment and opinion term triplets from sentences. Since the initial datasets used to evaluate models on ASTE had flaws, several studies later corrected the initial datasets and released new versions of the datasets independently. As a result, different studies select different versions of datasets to evaluate their methods, which makes ASTE-related works hard to follow. In this paper, we analyze the relation between different versions of datasets and suggest that the entire-space version should be used for ASTE. Besides the sentences containing triplets and the triplets in the sentences, the entire-space version additionally includes the sentences without triplets and the aspect terms which do not belong to any triplets. Hence, the entire-space version is consistent with real-world scenarios and evaluating models on the entire-space version can better reflect the models' performance in real-world scenarios. In addition, experimental results show that evaluating models on non-entire-space datasets inflates the performance of existing models and models trained on the entire-space version can obtain better performance~\footnote{Data and code are available at https://github.com/l294265421/entire-space-aste}.
\end{abstract}

\section{Introduction}
\begin{figure}
	\centering
	\includegraphics[scale=0.6]{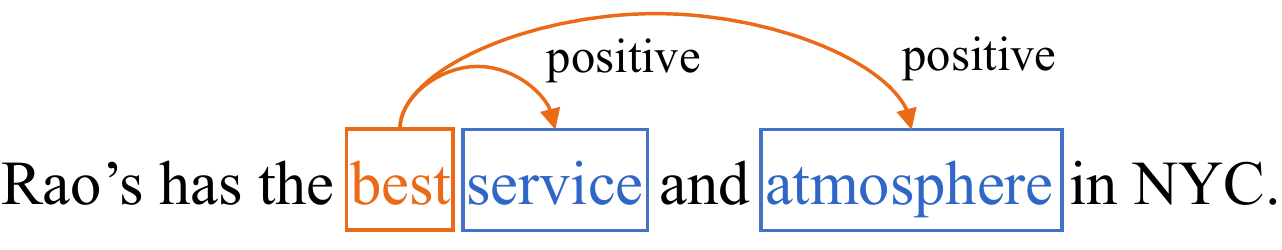}
	\caption{An example of ASTE. The words highlighted in blue represent aspect terms, while the words in orange represent the corresponding opinion terms.}
	\label{fig:an-example}
\end{figure}

Aspect Based Sentiment Analysis (ABSA)~\cite{liu2012sentiment,pontiki-etal-2014-semeval,pontiki-etal-2015-semeval,pontiki-etal-2016-semeval}, an important task in natural language understanding, receives much attention in recent years. Aspect Sentiment Triplet Extraction (ASTE) is a subtask of ABSA and is introduced by~\citet{Peng_Xu_Bing_Huang_Lu_Si_2020}. ASTE aims to extract (aspect term, sentiment, opinion term) triplets from sentences. Aspect terms are terms naming aspects. Opinion terms are words or phrases that express subjective attitudes explicitly. For example, given the sentence in Figure~\ref{fig:an-example}, ASTE extracts two triplets: (``service'', positive, ``best'') and (``atmosphere'', positive, ``best'').

To evaluate models on ASTE, five versions of ASTE datasets were constructed. The five versions of datasets are ASTE-Data-V1~\cite{Peng_Xu_Bing_Huang_Lu_Si_2020},  ASTE-Data-V2~\cite{xu-etal-2020-position}, ASTE-Data-GTS~\cite{wu-etal-2020-grid}~\footnote{GTS stands for Grid Tagging Scheme and is the model proposed for ASTE by \citet{wu-etal-2020-grid}.}, ASTE-Data-MTL~\cite{zhang-etal-2020-multi-task}~\footnote{MTL stands for Multi-task Learning and is the model proposed for ASTE by \citet{zhang-etal-2020-multi-task}.} and ASOTE-Data~\cite{li2021more}~\footnote{ASOTE-Data is introduced for the Aspect-Sentiment-Opinion Triplet Extraction (ASOTE) task by~\citet{li2021more}. The reason why ASOTE-Data is used for ASTE will be explained in Section~\ref{The-evolution-of-ASTE-Datasets}.}. All five versions of datasets derived from three SemEval tasks~\cite{pontiki-etal-2014-semeval,pontiki-etal-2015-semeval,pontiki-etal-2016-semeval} but are not the same. The relation between the five versions of datasets is described in section~\ref{The-evolution-of-ASTE-Datasets}. Here we only focus on one difference between these versions. While ASTE-Data-V1, ASTE-Data-V2, ASTE-Data-GTS and ASTE-Data-MTL only include the sentences with triplets and the triplets in these sentences, ASOTE-Data additionally include the sentences without triplets and the aspect terms that do not belong to any triplets. Table~\ref{comparision-of-different-versions} makes a comparison of the five versions of datasets by giving the annotations of four sentences in these datasets.

\begin{table*}\small
	\centering
	\begin{tabular}{|l|l|l|l|}
		\hline
		Sentences                                                                                   & ASTE-Data-V1                       & \begin{tabular}[c]{@{}l@{}}ASTE-Data-V2, \\ ASTE-Data-GTS, \\ ASTE-Data-MTL\end{tabular}                              & ASOTE-Data                                                                                                            \\ \hline
		\begin{tabular}[c]{@{}l@{}}Rao's has the best service\\ and atmosphere in NYC.\end{tabular} & (``service'', positive, ``best'')  & \begin{tabular}[c]{@{}l@{}}(``service'', positive, ``best''), \\ (``atmosphere'', positive, \\ ``best'')\end{tabular} & \begin{tabular}[c]{@{}l@{}}(``service'', positive, ``best''), \\ (``atmosphere'', positive, \\ ``best'')\end{tabular} \\ \hline
		\begin{tabular}[c]{@{}l@{}}Try the rose roll\\ (not on menu).\end{tabular}                  & (``rose roll'', positive, ``Try'') & (``rose roll'', positive, ``Try'')                                                                                    & \begin{tabular}[c]{@{}l@{}}(``rose roll'', positive, ``Try''), \\ (``menu'', , )\end{tabular}                         \\ \hline
		\begin{tabular}[c]{@{}l@{}}If you like spicy food get \\ the chicken vindaloo.\end{tabular} & do not include this sentence        & do not include this sentence                                                                                           & \begin{tabular}[c]{@{}l@{}}(``spicy food'', , ), \\ (``chicken vindaloo'', , )\end{tabular}                           \\ \hline
		\begin{tabular}[c]{@{}l@{}}It's always quiet \\ because it's awful.\end{tabular}            & do not include this sentence        & do not include this sentence                                                                                           &                                                                                                                       \\ \hline
	\end{tabular}
	\caption{\label{comparision-of-different-versions}
		A comparison of ASTE-Data-V1, ASTE-Data-V2, ASTE-Data-GTS, ASTE-Data-MTL and ASOTE-Data. All four sentences are from the Rest14 dataset in ASOTE-Data. In ASOTE-Data, the triplets only containing aspect terms are only used to train models and are not used to evaluate models. Sentences without triplets (e.g. the third and fourth sentences) will be used to evaluate models. The triplets only containing aspect terms are not regarded as triplets. Models probably wrongly recall triplets from sentences without triplets. 
	}
\end{table*}

However, researchers are not in agreement about which version of datasets should be used for ASTE. Different studies select different versions of datasets to train and evaluate their models and some studies even select more than one version of datasets. To name a few, \citet{mao2021joint}, \citet{chen2021bidirectional}, and \citet{chen-etal-2021-semantic} used ASTE-Data-V1. \citet{xu-etal-2021-learning}, \citet{mukherjee-etal-2021-paste}, and \citet{10.1145/3459637.3482058} used ASTE-Data-V2. \citet{yan-etal-2021-unified}, \citet{jing-etal-2021-seeking}, and \citet{liu-etal-2022-robustly} used ASTE-Data-V1 and ASTE-Data-V2. \citet{chen-etal-2022-enhanced} used ASTE-Data-V2 and ASTE-Data-GTS. \citet{9868116} used ASTE-Data-V2 and ASOTE-Data. In consequence, ASTE-related works are hard to follow. 

\begin{figure*}
	\centering
	\includegraphics[scale=0.44]{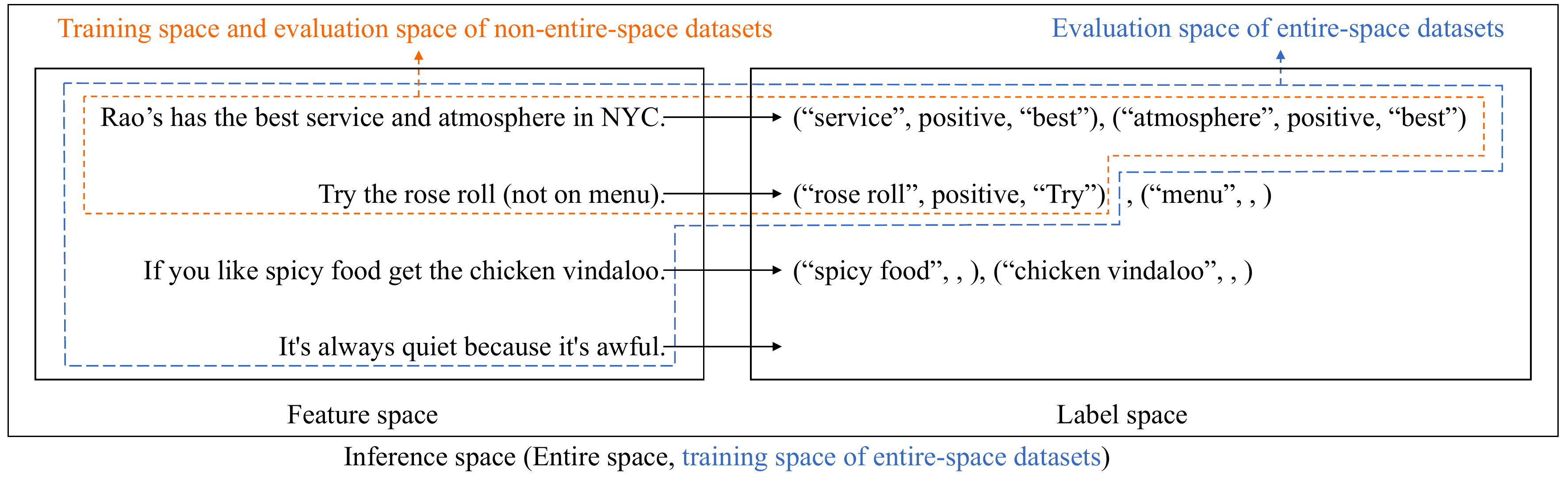}
	\caption{An illustration of the sample selection bias problem when non-entire-space datasets are used to train and evaluate ASTE models. Training space and evaluation space of non-entire-space datasets only include the sentences containing triplets. It is only part of the inference space which is composed of all sentences. Instead, entire space is the same as inference space. Entire space can be used to train ASTE models and entire space excluding the triplets only containing aspect terms should be used to evaluate ASTE models. The triplets only containing aspect terms are not regarded as triplets. Inference happens in real-world scenarios.}
	\label{fig:sample-selection-bias}
\end{figure*}

In this paper, we suggest that ASOTE-Data created by \citet{li2021more} should be used for ASTE. There are two reasons. First, as shown in Figure~\ref{fig:sample-selection-bias}, ignoring the sentences without triplets results in the sample selection bias problem: the training and evaluation space is not consistent with the inference space~\cite{zadrozny2004learning,5288526,ma2018entire,10.1145/3477495.3531768}. Specifically, the performance of ASTE models on the sentences with triplets cannot reflect their performance in real-world scenarios which include the sentences containing triplets and the sentences without triplets. Moreover, ASTE models are trained with only the sentences containing triplets, while these models will be used to make inference on both the sentences with triplets and the sentences without triplets. Thus, the generalization performance of trained models will be hurt. Second, the aspect terms that do not belong to any triplets can provide more supervisory signals to ASTE models. 

Since ASOTE-Data includes all types of sentences and annotations, we call it entire-space datasets. Datasets which only includes the sentences containing triplets and the triplets in these sentences, such as ASTE-Data-V1, ASTE-Data-V2, ASTE-Data-GTS and ASTE-Data-MTL, are non-entire-space datasets. Table~\ref{statistics-of-sentences-without-aspect-terms} show that there are a considerable amount of sentences without triplets and aspect terms that do not belong to any triplets in ASOTE-Data.

We conduct experiments to observe the differences between training and evaluating ASTE models on non-entire-space datasets and entire-space datasets. Results show that evaluating models on non-entire-space datasets inflates the performance of existing models and models trained on entire-space datasets can obtain better performance. 

It's worth noting that although \citet{li2021more} mentioned that ASOTE-Data is more appropriate than datasets only including the sentences with triplets to evaluate ASTE methods, they didn't analyze the differences between training and evaluating ASTE models on non-entire-space datasets and entire-space datasets. Thus, researchers may not have enough motivation to use ASOTE-Data. Moreover, when reading the code released by \citet{li2021more}, we find that the sentences without triplets and the aspect terms not in any triplets are utilized to train their model. But this is not mentioned in their paper. The possible reason is that \citet{li2021more} thought it is natural to train an ASTE model using all annotations during annotating ASTE datasets. However, influenced by many other ASTE studies, researchers, who do not know the details, will only use the sentences with triplets and the triplets in the sentences to train their proposed new models or other baselines and obtain worse results than the model of \citet{li2021more}, even if the new models and baselines are actually better. This also prevents researchers from using ASOTE-Data.

\section{The evolution of ASTE Datasets}
\label{The-evolution-of-ASTE-Datasets}

In this section, we first introduce the datasets used to construct ASTE datasets, then describe the evolution of ASTE datasets.

First, SemEval 2014 Task 4~\cite{pontiki-etal-2014-semeval} released Rest14-aspect-sentiment and Lap14-aspect-sentiment. SemEval 2015 task 12~\cite{pontiki-etal-2015-semeval} and SemEval 2016 task 5~\cite{pontiki-etal-2016-semeval} released Rest15-aspect-sentiment and Rest16-aspect-sentiment, respectively. The sentences in Rest14-aspect-sentiment, Rest15-aspect-sentiment and Rest16-aspect-sentiment are from the restaurant domain. Lap14-aspect-sentiment contains reviews from the laptop domain. The three  SemEval tasks only annotated aspect terms and their sentiments for the sentences in the four datasets. For example, for the sentence, ``those rolls were big, but not good and sashimi wasn't fresh.'', the aspect terms and their sentiments are annotated: (``rolls'', conflict) and (``sashimi'', negative).

Then, \citet{wang-etal-2016-recursive,wang2017coupled} annotated opinion terms and their sentiments for the sentences in Rest14-aspect-sentiment, Lap14-aspect-sentiment, and Rest15-aspect-sentiment, then obtained three new datasets: Rest14-opinion-sentiment, Lap14-opinion-sentiment, and Rest15-opinion-sentiment. For example, given the sentence, ``those rolls were big, but not good and sashimi wasn't fresh.'', the opinion terms and their sentiments are annotated: (``big'', positive), (``good'', positive) and (``fresh'', positive).

Later, \citet{fan-etal-2019-target} annotated  the corresponding opinion terms for the annotated aspect terms in Rest14-aspect-sentiment, Lap14-aspect-sentiment, Rest15-aspect-sentiment and Rest16-aspect-sentiment, then obtained four new datasets: Rest14-aspect-opinion-pair, Lap14-aspect-opinion-pair, Rest15-aspect-opinion-pair and Rest16-aspect-opinion-pair. For example, given the sentence, ``those rolls were big, but not good and sashimi wasn't fresh.'', the aspect term and opinion term pairs are annotated: (``rolls'', ``big''), (``rolls'', ``not good'') and (``sashimi'', ``wasn't fresh''). We can see that the guidelines of annotating opinion terms used by \citet{wang-etal-2016-recursive,wang2017coupled} and \citet{fan-etal-2019-target} are different.

\begin{figure*}
	\centering
	\includegraphics[scale=0.3]{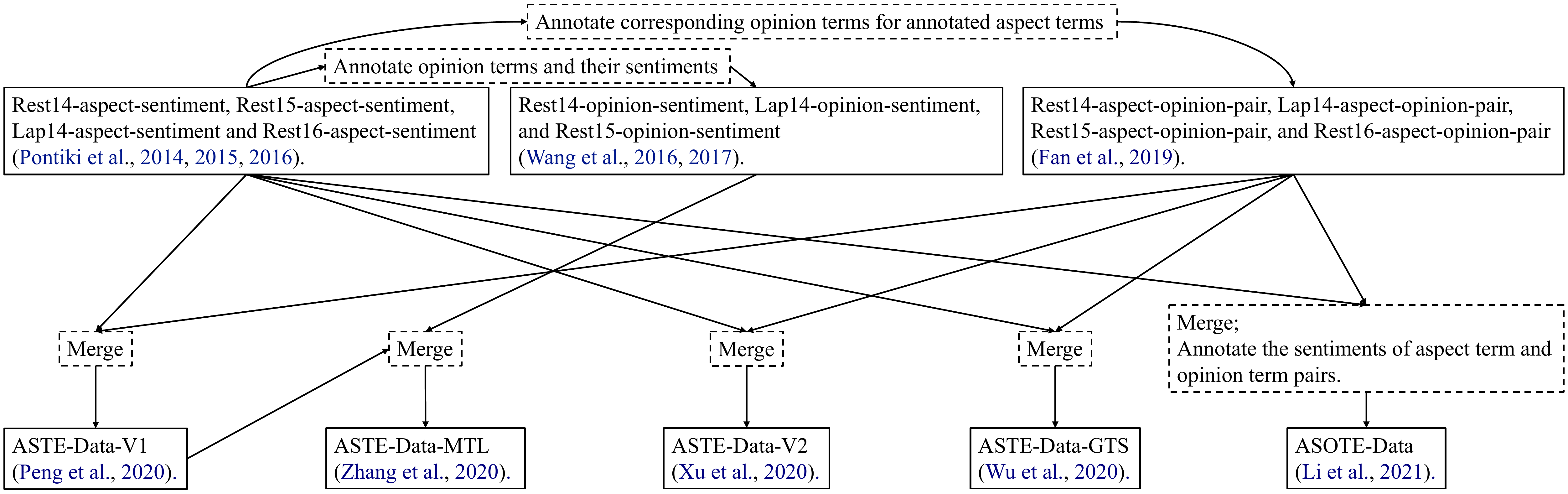}
	\caption{The relation between ASTE-related datasets.}
	\label{fig:evolution-of-datasets}
\end{figure*}

\begin{table*}\scriptsize
	\centering
	\begin{tabular}{cc|ccc|ccc|ccc|ccc|ccc}
		\hline
		\multicolumn{2}{c|}{\multirow{2}{*}{Datasets}} & \multicolumn{3}{c|}{ASTE-Data-V1} & \multicolumn{3}{c|}{ASTE-Data-V2} & \multicolumn{3}{c|}{ASTE-Data-GTS} & \multicolumn{3}{c|}{ASTE-Data-MTL} & \multicolumn{3}{c}{ASOTE-Data} \\ \cline{3-17} 
		\multicolumn{2}{c|}{}                          & \#S       & \#A       & \#T       & \#S       & \#A       & \#T       & \#S      & \#A      & \#T     & \#S      & \#A      & \#T     & \#S      & \#A      & \#T      \\ \hline
		\multirow{3}{*}{Rest14}         & Train        & 1299      & 2079      & 2145      & 1265      & 2051      & 2337      & 1259     & 2064     & 2356    & 1300     & 2077     & 2409    & 2429     & 2984     & 2499     \\
		& Dev          & 323       & 527       & 523       & 310       & 500       & 577       & 315      & 487      & 580     & 323      & 527      & 590     & 606      & 710      & 561      \\
		& Test         & 496       & 849       & 862       & 492       & 848       & 994       & 493      & 851      & 1008    & 496      & 849      & 1014    & 800      & 1134     & 1030     \\ \hline
		\multirow{3}{*}{Lap14}          & Train        & 917       & 1283      & 1265      & 900       & 1281      & 1460      & 899      & 1257     & 1452    & 920      & 1283     & 1451    & 2425     & 1927     & 1501     \\
		& Dev          & 226       & 317       & 337       & 219       & 296       & 345       & 225      & 332      & 383     & 228      & 315      & 380     & 608      & 437      & 347      \\
		& Test         & 339       & 475       & 490       & 328       & 463       & 541       & 332      & 467      & 547     & 339      & 474      & 552     & 800      & 655      & 563      \\ \hline
		\multirow{3}{*}{Rest15}         & Train        & 593       & 834       & 923       & 605       & 862       & 1013      & 603      & 871      & 1038    & 593      & 834      & 977     & 1050     & 950      & 1031     \\
		& Dev          & 148       & 225       & 238       & 148       & 213       & 249       & 151      & 205      & 239     & 148      & 225      & 260     & 263      & 249      & 246      \\
		& Test         & 318       & 426       & 455       & 322       & 432       & 485       & 325      & 436      & 493     & 318      & 426      & 479     & 684      & 542      & 493      \\ \hline
		\multirow{3}{*}{Rest16}         & Train        & 842       & 1183      & 1289      & 857       & 1198      & 1394      & 863      & 1213     & 1421    & 842      & 1183     & 1370    & 1595     & 1399     & 1431     \\
		& Dev          & 210       & 291       & 316       & 210       & 296       & 339       & 216      & 298      & 348     & 210      & 291      & 334     & 400      & 344      & 333      \\
		& Test         & 319       & 444       & 465       & 325       & 452       & 514       & 328      & 456      & 525     & 320      & 444      & 507     & 675      & 612      & 524      \\ \hline
	\end{tabular}
	\caption{\label{statistics-of-different-versions}
		Statistics of ASTE-Data-V1, ASTE-Data-V2, ASTE-Data-GTS, ASTE-Data-MTL and ASOTE-Data. Here ``\#S'', ``\#A'', and ``\#T'' denote the numbers of sentences, aspect terms and triplets, respectively.
	}
\end{table*}

\begin{table}\scriptsize
	\centering
	\begin{tabular}{cc|ccc|cc}
		\hline
		\multicolumn{2}{c|}{\multirow{2}{*}{Datasets}} & \multicolumn{3}{c|}{\begin{tabular}[c]{@{}c@{}}sentences\\ \_with\_triplets\end{tabular}} & \multicolumn{2}{c}{\begin{tabular}[c]{@{}c@{}}sentences\\ \_without\_triplets\end{tabular}} \\ \cline{3-7} 
		\multicolumn{2}{c|}{}                          & \#S                          & \#A1                         & \#A2                        & \#S                                          & \#A2                                         \\ \hline
		\multirow{3}{*}{Rest14}         & Train        & 1301                         & 2138                         & 342                         & 1128                                         & 504                                          \\
		& Dev          & 323                          & 500                          & 73                          & 283                                          & 137                                          \\
		& Test         & 500                          & 865                          & 92                          & 300                                          & 177                                          \\ \hline
		\multirow{3}{*}{Lap14}          & Train        & 917                          & 1304                         & 205                         & 1508                                         & 418                                          \\
		& Dev          & 226                          & 305                          & 32                          & 382                                          & 100                                          \\
		& Test         & 342                          & 480                          & 53                          & 458                                          & 122                                          \\ \hline
		\multirow{3}{*}{Rest15}         & Train        & 611                          & 864                          & 22                          & 439                                          & 64                                           \\
		& Dev          & 143                          & 212                          & 11                          & 120                                          & 26                                           \\
		& Test         & 325                          & 436                          & 17                          & 359                                          & 89                                           \\ \hline
		\multirow{3}{*}{Rest16}         & Train        & 862                          & 1218                         & 35                          & 733                                          & 146                                          \\
		& Dev          & 211                          & 289                          & 15                          & 189                                          & 40                                           \\
		& Test         & 328                          & 456                          & 61                          & 347                                          & 95                                           \\ \hline
	\end{tabular}
	\caption{\label{statistics-of-sentences-without-aspect-terms}
		Some statistics of ASOTE-Data. Here ``\#S'', ``\#A1'', and ``\#A2'' denote the numbers of sentences, aspect terms in triplets and aspect terms that do not belong to any triplets, respectively.
	}
\end{table}

\citet{Peng_Xu_Bing_Huang_Lu_Si_2020} created four datasets (i.e. Rest14, Lap14, Rest15 and Rest16) for ASTE by combining the four datasets (i.e. Rest14-aspect-sentiment, Lap-aspect-sentiment, Rest15-aspect-sentiment and Rest16-aspect-sentiment) released by the three  SemEval tasks with the four datasets annotated by \citet{fan-etal-2019-target} (i.e.  Rest14-aspect-opinion-pair, Lap14-aspect-opinion-pair, Rest15-aspect-opinion-pair and Rest16-aspect-opinion-pair). This version of ASTE datasets is called ASTE-Data-V1. ASTE-Data-V1 has a flaw. If an opinion term is associated with multiple aspect terms, only one aspect term among these aspect terms  and the opinion term are annotated as one triplet. For example, for the sentence ``Rao’s has the best service and atmosphere in NYC.
'', since the two aspect terms ``service'' and ``atmosphere'' share the opinion term ``best'', only the triplet (``service'', positive, ``best'') is annotated.

\citet{xu-etal-2020-position}, \citet{wu-etal-2020-grid}, and \citet{zhang-etal-2020-multi-task} corrected ASTE-Data-V1 and released three new versions of datasets. Similar to \citet{Peng_Xu_Bing_Huang_Lu_Si_2020}, \citet{xu-etal-2020-position} and \citet{wu-etal-2020-grid} aligned the four datasets released by the three  SemEval tasks and the four datasets annotated by \citet{fan-etal-2019-target}, and built two new versions of datasets, called ASTE-Data-V2 and ASTE-Data-GTS respectively. \citet{zhang-etal-2020-multi-task} corrected ASTE-Data-V1 by aligning ASTE-Data-V1 with the datasets (i.e. Rest14-opinion-sentiment, Lap14-opinion-sentiment, and Rest15-opinion-sentiment) created by~\citet{wang-etal-2016-recursive,wang2017coupled} and created a new version of datasets called ASTE-Data-MTL. Note that, the triplets with ``conflict'' sentiments are not included by ASTE-Data-V1, ASTE-Data-V2, ASTE-Data-GTS and ASTE-Data-MTL.

\begin{figure}
	\centering
	\includegraphics[scale=0.45]{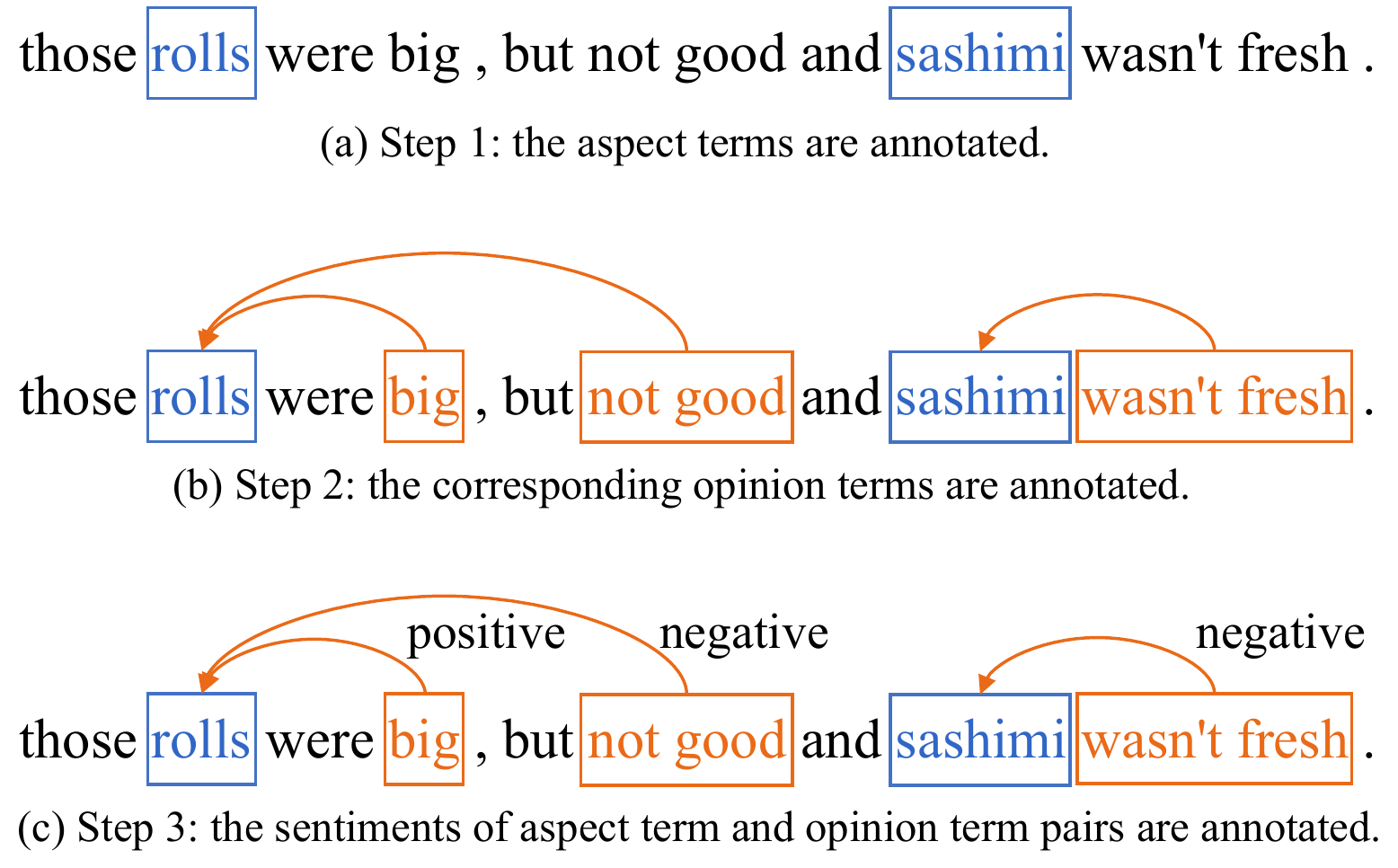}
	\caption{The annotation process of ASOTE-Data. ASOTE-Data includes all annotations produced during constructing the datasets. The aspect terms that do not belong to any triplets are not additionally annotated.}
	\label{fig:the-asote-data-annotation-process}
\end{figure}

ASTE-Data-V1, ASTE-Data-V2, ASTE-Data-GTS and ASTE-Data-MTL only include the sentences containing triplets and the triplets in these sentences. Moreover, the triplets with ``conflict'' sentiments are also removed from these datasets. Thus, these datasets are not consistent with real-world scenarios. Therefore, \citet{li2021more} constructed a new version of datasets by aligning the four datasets released by the three  SemEval tasks and the four datasets annotated by \citet{fan-etal-2019-target} and keeping both the sentences containing triplets and the sentences without triplets. Moreover, \citet{li2021more} replaced the  sentiments of aspect terms with the sentiments of aspect term-opinion term pairs. Aspect term and opinion term pairs do not have ``conflict'' sentiments. What is more, the aspect terms which do not belong to any triplets are included to train models. This version of datasets is called ASOTE-Data. We can see that ASTE-Data-V1, ASTE-Data-V2, ASTE-Data-GTS and ASTE-Data-MTL are subsets of ASOTE-Data. In summary, ASOTE-Data is annotated through three steps. First, the aspect terms in sentences are annotated. Second, for the annotated aspect terms, their corresponding opinion terms are annotated. Finally, the sentiments of the annotated aspect term and opinion term pairs are annotated. The annotation process is illustrated in Figure~\ref{fig:the-asote-data-annotation-process}. 

The relation between the datasets mentioned above is shown in Figure~\ref{fig:evolution-of-datasets}. Some statistics of the five versions of datasets are shown in Table~\ref{statistics-of-different-versions}.

It's worth mentioning that ASOTE-Data was built for Aspect-Sentiment-Opinion Triplet Extraction (ASOTE) Task~\cite{li2021more} rather than ASTE. Similar to ASTE, ASOTE also extracts (aspect term, sentiment, opinion term) triplets from sentences. There is only one difference between ASTE and ASOTE. While the sentiment in a triplet extracted by ASTE is the sentiment that the sentence expresses toward the aspect term, the sentiment in a triplet extracted by ASOTE is the sentiment of the aspect term and opinion term pair. For example, given the sentence ``The atmosphere is attractive, but a little uncomfortable.'', ASTE extracts (“atmosphere”, conflict, “attractive”) and (“atmosphere”, conflict, “uncomfortable”). And ASOTE extracts (“atmosphere”, positive, “attractive”) and (“atmosphere”, negative, “uncomfortable”). However, all ASTE datasets (i.e. ASTE-Data-V1, ASTE-Data-V2, ASTE-Data-GTS and ASTE-Data-MTL) removes the triplets with ``conflict'' sentiment. In addition, the triplets with ``conflict'' sentiments are not many and the sentiments of most aspect term and opinion term pairs in ASOTE-Data are the same as the sentiments of the aspects whose sentiments are not ``conflict''. Therefore, the ASTE task that existing works attempt to solve is the same as ASOTE.

\begin{table*}\scriptsize
	\centering
	\begin{tabular}{c|c|ccc|ccc|ccc|ccc}
		\hline
		\multirow{2}{*}{Method}      & \multirow{2}{*}{Test} & \multicolumn{3}{c|}{Rest14}    & \multicolumn{3}{c|}{Lap14}     & \multicolumn{3}{c|}{Rest15}    & \multicolumn{3}{c}{Rest16}     \\ \cline{3-14} 
		&                       & P     & R     & F1             & P     & R     & F1             & P     & R     & F1             & P     & R     & F1             \\ \hline
		\multirow{3}{*}{GTS-BiLSTM}  & Entire-Space          & 68.50 & 52.01 & 59.01          & 55.11 & 33.67 & 41.66          & 57.34 & 45.85 & 50.94          & 55.57 & 57.22 & 55.78          \\ \cline{2-14} 
		& Non-Entire-Space      & 74.09 & 52.01 & 61.03          & 63.52 & 33.67 & 43.92          & 70.37 & 45.85 & 55.50          & 68.95 & 57.22 & 62.00          \\
		& (Gains (\%))          & 7.55  & 0.00  & 3.30           & 13.24 & 0.00  & 5.13           & 18.51 & 0.00  & 8.22           & 19.41 & 0.00  & 10.02          \\ \hline
		\multirow{3}{*}{GTS-CNN}     & Entire-Space          & 64.64 & 60.28 & 62.38          & 42.88 & 42.46 & 42.65          & 48.85 & 51.45 & 49.96          & 52.96 & 61.00 & 56.67          \\ \cline{2-14} 
		& Non-Entire-Space      & 71.44 & 60.28 & 65.39          & 53.04 & 42.46 & 47.13          & 60.17 & 51.45 & 55.35          & 65.11 & 61.00 & 62.97          \\
		& (Gains (\%))          & 9.52  & 0.00  & 4.60           & 19.16 & 0.00  & 9.52           & 18.81 & 0.00  & 9.74           & 18.67 & 0.00  & 10.02          \\ \hline
		\multirow{3}{*}{Span-BiLSTM} & Entire-Space          & 68.48 & 62.47 & 65.32          & 51.98 & 47.03 & 49.37          & 56.22 & 53.63 & 54.88          & 57.25 & 65.57 & 61.11          \\ \cline{2-14} 
		& Non-Entire-Space      & 74.56 & 62.47 & 67.97          & 61.30 & 47.03 & 53.22          & 66.84 & 53.63 & 59.49          & 68.57 & 65.57 & 67.03          \\
		& (Gains (\%))          & 8.15  & 0.00  & 3.89           & 15.21 & 0.00  & 7.22           & 15.89 & 0.00  & 7.76           & 16.50 & 0.00  & 8.83           \\ \hline
		\multirow{3}{*}{GTS-BERT}    & Entire-Space          & 62.07 & 67.60 & 64.69          & 46.39 & 51.74 & 48.79          & 51.79 & 58.00 & 54.64          & 56.02 & 68.19 & 61.47          \\ \cline{2-14} 
		& Non-Entire-Space      & 69.82 & 67.60 & 68.66          & 57.64 & 51.74 & 54.48          & 62.13 & 58.00 & 59.94          & 69.35 & 68.19 & 68.73          \\
		& (Gains (\%))          & 11.11 & 0.00  & 5.78           & 19.52 & 0.00  & 10.44          & 16.64 & 0.00  & 8.84           & 19.22 & 0.00  & 10.56          \\ \hline
		\multirow{3}{*}{PBF-BERT}    & Entire-Space          & 56.29 & 73.55 & 63.73          & 39.51 & 61.39 & 48.03          & 43.88 & 64.54 & 52.10          & 46.16 & 72.67 & 56.37          \\ \cline{2-14} 
		& Non-Entire-Space      & 65.52 & 73.55 & 69.28          & 52.29 & 61.39 & 56.44          & 58.07 & 64.54 & 61.06          & 63.37 & 72.67 & 67.65          \\
		& (Gains (\%))          & 14.08 & 0.00  & 8.01           & 24.43 & 0.00  & 14.89          & 24.43 & 0.00  & 14.68          & 27.15 & 0.00  & 16.68          \\ \hline
		\multirow{3}{*}{Span-BERT}   & Entire-Space          & 65.62 & 70.76 & \textbf{68.08} & 48.55 & 58.97 & \textbf{53.23} & 50.40 & 65.36 & \textbf{56.91} & 55.23 & 73.13 & \textbf{62.87} \\ \cline{2-14} 
		& Non-Entire-Space      & 74.40 & 70.76 & {\ul 72.52}    & 61.62 & 58.97 & {\ul 60.24}    & 62.52 & 65.36 & {\ul 63.90}    & 71.13 & 73.13 & {\ul 72.08}    \\
		& (Gains (\%))          & 11.81 & 0.00  & 6.13           & 21.21 & 0.00  & 11.63          & 19.38 & 0.00  & 10.94          & 22.36 & 0.00  & 12.78          \\ \hline
	\end{tabular}
	\caption{\label{Evaluation-on-Entire-Space}
		Performance of the ASTE models evaluated on entire-space and non-entire-space test sets. All models are trained on non-entire-space data.	Gains indicate how much higher the performance of the models evaluated on non-entire-space datasets is than the performance of the models evaluated on entire-space datasets. The best F1 scores on entire-space datasets are marked in bold and the best F1 scores on non-entire-space datasets are underlined.
	}
\end{table*}

\section{Related Work}
\textbf{Aspect Sentiment Triplet Extraction} (ASTE) is a popular subtask of Aspect Based Sentiment Analysis (ABSA) in recent years. Many models have been proposed for ASTE. \citet{wu-etal-2020-grid} proposed Grid Tagging Scheme (GTS) to address ASTE. While GTS rely on the interactions between each aspect word and opinion word, \citet{xu-etal-2021-learning} proposed a span-level approach which considers the interaction between the whole spans of aspect terms and opinion terms. \citet{chen-etal-2022-enhanced} defined ten types of relations between words and proposed an Enhanced Multi-Channel Graph Convolutional Network model (EMCGCN) for ASTE to fully utilize the relations between words. However, these works use different versions of ASTE datasets and hence make ASTE-related works hard to follow. In addition, except for \citet{li2021more,9868116}, all studies used non-entire-space datasets to evaluate the performance of their methods and hence the performance of their methods in real-world scenarios is not reflected.

\textbf{Sample Selection Bias} has been studied in other domains~\cite{zadrozny2004learning,ma2018entire,10.1145/3477495.3531768}. The work of~\citet{10.1145/3477495.3531768} is the one most related to ours. They studied the sample selection bias of Target-oriented Opinion Words Extraction (TOWE) models trained on non-entire-space datasets. There are two key differences between their work and ours. First, the aim of this paper is not to study the sample selection bias problem of ASTE, instead, to select a proper version of datasets among five versions of datasets for ASTE. Second, TOWE is a subtask of ASTE. Besides TOWE, another subtask of ASTE, Aspect Term Extraction, also has the sample selection bias problem, which is explored in this paper.

\section{Experiments}

\subsection{Datasets and Metrics}
To observe the differences between training and evaluating ASTE models on non-entire-space datasets and entire-space datasets, we conducted experiments on ASOTE-Data. The non-entire-space version of ASOTE-Data only includes the sentences containing triplets and the triplets in these sentences. Table~\ref{statistics-of-different-versions} and Table~\ref{statistics-of-sentences-without-aspect-terms} present some key statistics of ASOTE-Data.

We use precision (P), recall (R), and F1-score (F1) as the evaluation metrics. For the ASTE task, an extracted triplet is regarded as correct only if the predicted aspect term span, sentiment, opinion term span and ground truth aspect term span, sentiment, opinion term span are exactly matched. We also explore the aspect term extraction task. An extracted aspect term is regarded as correct only if the predicted aspect term span and the ground truth aspect term span is exactly matched.

\subsection{Models}
We run experiments based on six ASTE models. The six models are i) five end-to-end models: GTS-LiLSTM, GTS-CNN, GTS-BERT~\citet{wu-etal-2020-grid}\footnote{https://github.com/NJUNLP/GTS}, Span-BiLSTM, Span-BERT~\cite{xu-etal-2021-learning}\footnote{https://github.com/chiayewken/Span-ASTE}, and ii) one pipeline method: PBF-BERT~\cite{li2021more}\footnote{https://github.com/l294265421/ASOTE}. The suffixes BiLSTM, CNN and BERT of these model names indicate the sentence encoder of theses models.

Apart from PBF-BERT, the released code of all other five models do not support that the models are trained and evaluated on entire space. Thus, we adjust their code and release the code. We run all models for 5 times and report the average results on the test datasets.

\subsection{Evaluation on Entire Space}
To observe the differences between evaluating ASTE models on entire-space datasets and non-entire-space datasets, the six ASTE models are trained on non-entire-space datasets (Both the training set and development set are non-entire-space) like most previous studies~\cite{xu-etal-2020-position,xu-etal-2021-learning}, then are evaluated on entire-space datasets. Table~\ref{Evaluation-on-Entire-Space} presents the experimental results. We have two observations. First, the performance of the models on non-entire-space datasets across all four datasets is much higher than their performance on entire-space datasets in terms of F1 scores. For example, The performance gains of the best model Span-BERT on Rest14, Lap14, Rest15 and Rest16 are 6.13\%, 11.63\%, 10.94\% and 12.78 \%, respectively. That is to say, evaluating ASTE models on non-entire-space datasets inflates model performance and hence cannot reflect the performance of ASTE models in real-world scenarios. Second, the recall scores of ASTE models on entire-space datasets and non-entire-space datasets are the same. The reason is that all correct triplets are included in the non-entire-space datasets and hence the ASTE models cannot recall more correct triplets from the sentences without triplets. In addition, the precision scores of models on entire space are smaller than that on non-entire-space datasets. The reason is that the models wrongly recall triplets from the sentences without triplets. 

\begin{table*}\scriptsize
	\centering
	\begin{tabular}{c|c|ccc|ccc|ccc|ccc}
		\hline
		\multirow{2}{*}{Method}      & \multirow{2}{*}{Train-Dev} & \multicolumn{3}{c|}{Rest14}    & \multicolumn{3}{c|}{Lap14}      & \multicolumn{3}{c|}{Rest15}    & \multicolumn{3}{c}{Rest16}     \\ \cline{3-14} 
		&                            & P     & R     & F1             & P     & R      & F1             & P     & R     & F1             & P     & R     & F1             \\ \hline
		\multirow{3}{*}{GTS-BiLSTM}  & Non-Entire-Space           & 68.50 & 52.01 & 59.01          & 55.11 & 33.67  & 41.66          & 57.34 & 45.85 & 50.94          & 55.57 & 57.22 & 55.78          \\ \cline{2-14} 
		& Entire-Space               & 71.23 & 56.42 & 62.96          & 57.08 & 34.98  & 43.34          & 69.36 & 43.85 & 53.69          & 64.11 & 57.92 & 60.84          \\
		& (Gains (\%))               & 3.99  & 8.48  & 6.69           & 3.58  & 3.91   & 4.03           & 20.96 & -4.37 & 5.39           & 15.37 & 1.21  & 9.06           \\ \hline
		\multirow{3}{*}{GTS-CNN}     & Non-Entire-Space           & 64.64 & 60.28 & 62.38          & 42.88 & 42.46  & 42.65          & 48.85 & 51.45 & 49.96          & 52.96 & 61.00 & 56.67          \\ \cline{2-14} 
		& Entire-Space               & 69.97 & 57.88 & 63.33          & 55.32 & 39.40  & 45.94          & 60.68 & 48.39 & 53.75          & 58.65 & 63.01 & 60.63          \\
		& (Gains (\%))               & 8.25  & -3.97 & 1.53           & 29.01 & -7.21  & 7.72           & 24.21 & -5.96 & 7.59           & 10.75 & 3.29  & 6.99           \\ \hline
		\multirow{3}{*}{Span-BiLSTM} & Non-Entire-Space           & 68.48 & 62.47 & 65.32          & 51.98 & 47.03  & 49.37          & 56.22 & 53.63 & 54.88          & 57.25 & 65.57 & 61.11          \\ \cline{2-14} 
		& Entire-Space               & 73.89 & 58.70 & 65.38          & 64.23 & 42.66  & 51.25          & 64.23 & 50.59 & 56.57          & 65.93 & 60.31 & 62.95          \\
		& (Gains (\%))               & 7.90  & -6.03 & 0.09           & 23.57 & -9.29  & 3.80           & 14.24 & -5.67 & 3.07           & 15.16 & -8.03 & 3.00           \\ \hline
		\multirow{3}{*}{GTS-BERT}    & Non-Entire-Space           & 62.07 & 67.60 & 64.69          & 46.39 & 51.74  & 48.79          & 51.79 & 58.00 & 54.64          & 56.02 & 68.19 & 61.47          \\ \cline{2-14} 
		& Entire-Space               & 69.33 & 66.41 & 67.83          & 58.71 & 49.11  & 53.46          & 61.45 & 54.52 & 57.75          & 60.01 & 68.30 & 63.87          \\
		& (Gains (\%))               & 11.70 & -1.75 & 4.85           & 26.57 & -5.09  & 9.56           & 18.65 & -5.99 & 5.69           & 7.11  & 0.17  & 3.89           \\ \hline
		\multirow{3}{*}{PBF-BERT}    & Non-Entire-Space           & 56.29 & 73.55 & 63.73          & 39.51 & 61.39  & 48.03          & 43.88 & 64.54 & 52.10          & 46.16 & 72.67 & 56.37          \\ \cline{2-14} 
		& Entire-Space               & 67.68 & 69.73 & {\ul 68.69}    & 54.58 & 56.48  & \textbf{55.49} & 56.79 & 61.79 & {\ul 59.13}    & 61.10 & 72.52 & {\ul 66.31}    \\
		& (Gains (\%))               & 20.23 & -5.20 & 7.77           & 38.13 & -7.99  & 15.51          & 29.41 & -4.27 & 13.51          & 32.35 & -0.21 & 17.63          \\ \hline
		\multirow{3}{*}{Span-BERT}   & Non-Entire-Space           & 65.62 & 70.76 & 68.08          & 48.55 & 58.97  & 53.23          & 50.40 & 65.36 & 56.91          & 55.23 & 73.13 & 62.87          \\ \cline{2-14} 
		& Entire-Space               & 73.34 & 66.56 & \textbf{69.71} & 61.37 & 50.48  & {\ul 55.33}    & 58.71 & 61.91 & \textbf{60.21} & 64.70 & 68.74 & \textbf{66.64} \\
		& (Gains (\%))               & 11.77 & -5.93 & 2.40           & 26.41 & -14.40 & 3.94           & 16.48 & -5.28 & 5.81           & 17.16 & -6.00 & 6.01           \\ \hline
	\end{tabular}
	\caption{\label{Training-on-Entire-Space}
		Performance of the ASTE models trained on non-entire-space and entire-space datasets. All models are evaluated on entire-space datasets. Gains indicate how much better the model trained on entire-space datasets is than the model trained on non-entire-space datasets. The best F1 scores are marked in bold and the second best F1 scores are underlined.
	}
\end{table*}

\subsection{Training on Entire Space}
In this section, we explore the differences between training ASTE models on non-entire-space and entire-space datasets. Specifically, the six ASTE models are trained in two settings: i) both the training set and development set are non-entire-space, and ii) both the training set and development set are entire-space. Then the trained models are evaluated on entire-space datasets. Experimental results are shown in Table~\ref{Training-on-Entire-Space}. From Table~\ref{Training-on-Entire-Space} we draw the following two conclusions. First, all models trained on entire-space datasets surpass them trained on non-entire-space datasets in terms of F1 score across all four datasets, indicating that training models on entire-space datasets can improve the generalization performance of ASTE models. Second, the pipeline method PBF-BERT trained on entire-space datasets outperforms all other models trained on non-entire-space datasets on all four datasets. Thus, when a new model is proposed and is compared with PBF-BERT on ASOTE-Data, training the new model on entire space is critical. 

\subsection{Case Study}
To intuitively understand the differences between training ASTE models on non-entire-space datasets and entire-space datasets, we show the predictions of the best model Span-BERT trained on non-entire-space datasets and entire-space datasets on one sentence without triplets. Given the sentence, ``However, it was worth the visit.'',  in the test set of Rest14, while Span-BERT trained on non-entire-space datasets extracts one triplet: (``visit'', positive, ``worth''), Span-BERT trained on entire space extracts nothing. One possible reason is that the training set of Rest14 has a similar sentence without triplets: ``Worth a visit.''.

\begin{figure}
	\centering
	\includegraphics[scale=0.5]{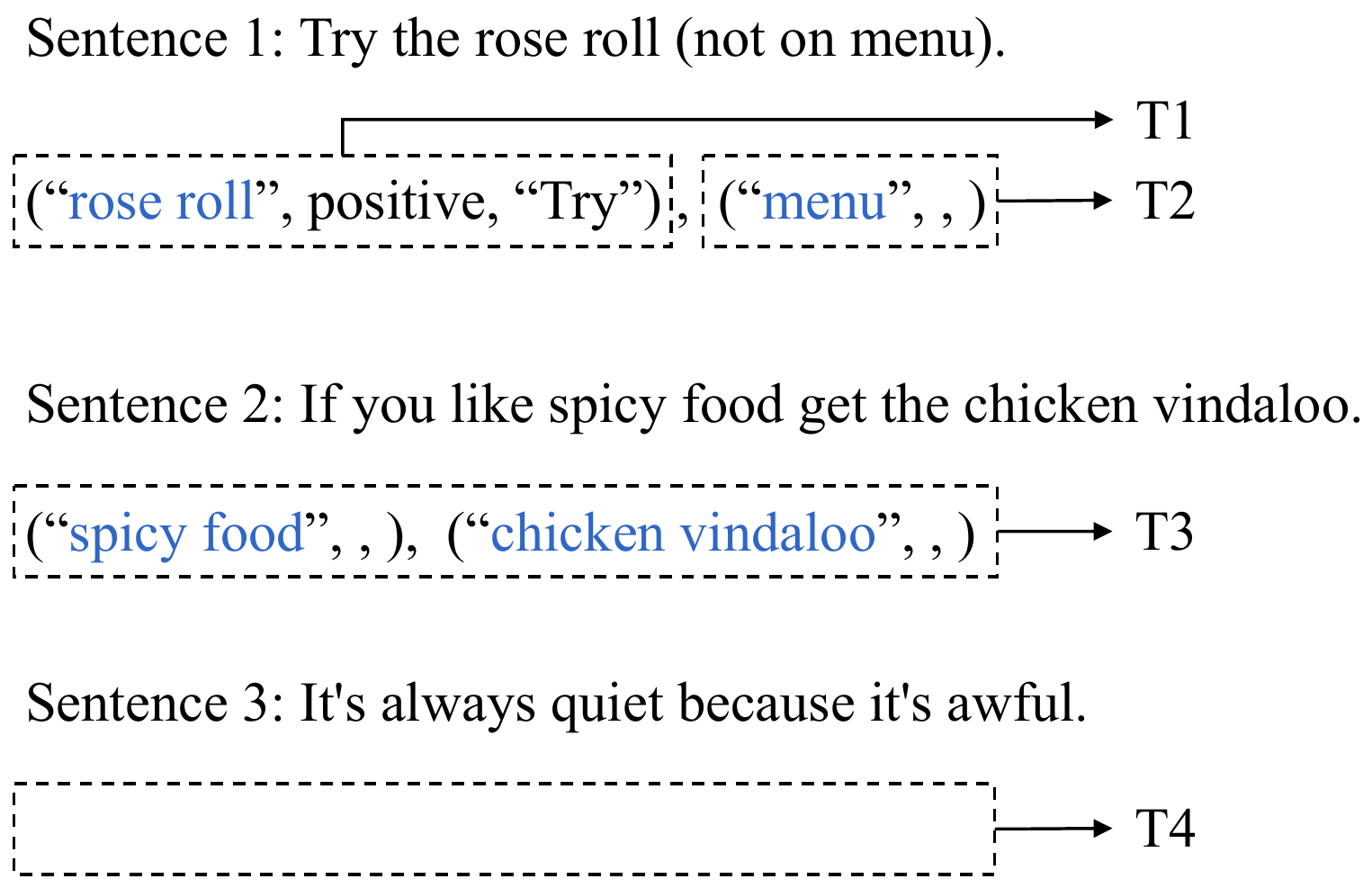}
	\caption{The four types of ATE instances.}
	\label{fig:ATE-instance-types}
\end{figure}

\begin{table*}\scriptsize
	\centering
	\begin{tabular}{c|c|ccc|ccc|ccc|ccc}
		\hline
		\multirow{2}{*}{Method}   & \multirow{2}{*}{Train-Dev} & \multicolumn{3}{c|}{Rest14}                      & \multicolumn{3}{c|}{Lap14}                       & \multicolumn{3}{c|}{Rest15}                      & \multicolumn{3}{c}{Rest16}                       \\ \cline{3-14} 
		&                            & P              & R              & F1             & P              & R              & F1             & P              & R              & F1             & P              & R              & F1             \\ \hline
		\multirow{4}{*}{ATE-BERT} & T1                         & 82.77          & 82.31          & 82.49          & 69.02          & 77.68          & 73.04          & 58.34          & 77.05          & 66.34          & 65.68          & 81.11          & 72.58          \\
		& T1+T2                      & 82.04          & 88.10          & 84.96          & 70.79          & 83.30          & 76.51          & 58.35          & 76.46          & 66.16          & 63.53          & 83.34          & 72.06          \\
		& T1+T2+T3                   & 82.31          & \textbf{89.15} & 85.58          & 71.23          & \textbf{83.82} & 76.93          & 59.65          & \textbf{80.55} & 68.49          & 66.44          & \textbf{85.34} & 74.69          \\ \cline{2-14} 
		& Entire Space               & \textbf{87.01} & 88.36          & \textbf{87.68} & \textbf{82.72} & 81.35          & \textbf{82.01} & \textbf{68.36} & 72.95          & \textbf{70.55} & \textbf{75.33} & 76.53          & \textbf{75.91} \\ \hline
	\end{tabular}
	\caption{\label{ATE-training}
		Performance of ATE-BERT trained on different combinations of ATE instance types. ATE-BERT is evaluated on entire space. The best scores are marked in bold.
	}
\end{table*}

\begin{table*}\scriptsize
	\centering
	\begin{tabular}{c|c|ccc|ccc|ccc|ccc}
		\hline
		\multirow{2}{*}{Method}   & \multirow{2}{*}{Test} & \multicolumn{3}{c|}{Rest14}                      & \multicolumn{3}{c|}{Lap14}                       & \multicolumn{3}{c|}{Rest15}                      & \multicolumn{3}{c}{Rest16}                       \\ \cline{3-14} 
		&                       & P              & R              & F1             & P              & R              & F1             & P              & R              & F1             & P              & R              & F1             \\ \hline
		\multirow{2}{*}{ATE-BERT} & Entire Space          & 82.77          & 82.31          & 82.49          & 69.02          & 77.68          & 73.04          & 58.34          & 77.05          & 66.34          & 65.68          & 81.11          & 72.58          \\
		& T1                    & \textbf{83.66} & \textbf{87.19} & \textbf{85.38} & \textbf{78.37} & \textbf{82.05} & \textbf{80.16} & \textbf{76.22} & \textbf{81.33} & \textbf{78.67} & \textbf{75.15} & \textbf{85.44} & \textbf{79.96} \\ \hline
	\end{tabular}
	\caption{\label{ATE-evaluation}
		Performance of ATE-BERT on entire space and T1. ATE-BERT is trained on T1.
	}
\end{table*}

\subsection{Subtasks}
To further understand the differences between training and evaluating models on non-entire-space datasets and entire-space datasets, we decompose ASTE into three subtasks and analyze the subtasks. Specifically, the three subtasks are Aspect Term Extraction (ATE)~\cite{pontiki-etal-2014-semeval}, Target-oriented Opinion Words Extraction (TOWE)~\cite{fan-etal-2019-target} and Aspect-Opinion Pair Sentiment Classification (AOPSC)~\cite{li2021more}. ATE extracts aspect terms from sentences, TOWE extracts corresponding opinion terms for the aspect terms detected by ATE, and AOPSC predicts the sentiment of a given aspect term and opinion term pair. The process of using the three subtasks to extract triplets from sentences is similar to the annotation process of ASOTE-Data presented in Figure~\ref{fig:the-asote-data-annotation-process}. We find that AOPSC has the same training and evaluation data in non-entire-space and entire-space datasets, hence we only analyze ATE and TOWE.

\subsubsection{ATE Task}

We run experiments using the Aspect Term Extraction (ATE) model in the pipeline ASTE model PBF-BERT. The ATE model is named ATE-BERT. 

We first explore the differences between training ATE-BERT on different types of ATE instances. As shown in Figure~\ref{fig:ATE-instance-types}, there exist four types of ATE instances:
\begin{itemize}
	\item T1: aspect terms in triplets.
	\item T2: aspect terms which do not belong to any triplets but appear in the sentences containing triplets.
	\item T3: aspect terms which do not belong to any triplets and appear in the sentences without triplets.
	\item T4: sentences that do not contain aspect terms.
\end{itemize}

Non-entire-space ASTE datasets only include T1. We train ATE-BERT on different combinations of these types of instances and evaluate model on entire space including all types of instances. Experimental results are presented in Table~\ref{ATE-training}. We have two observations about the experimental results. First, compared with ATE-BERT trained only on T1, ATE-BERT trained on T1 and T2 obtains better performance on Rest14 and Lap14 and achieves similar performance on Rest15 and Rest16 in terms of F1 scores. By additionally including T3, the performance of ATE-BERT is further improved. This indicates that the annotated aspect terms which do not belong to any triplets should be used to train ASTE models. Second, ATE-BERT trained on entire space obtains the best F1 scores on all four datasets because of the sentences that do not contain aspect terms (i.e. T4). The reason is that the sentences without aspect terms in training sets prevent ATE-BERT wrongly recalling aspect terms from sentences without aspect terms in test sets. The evidence is that ATE-BERT trained on entire space obtains worse recall scores than ATE-BERT trained on T1, T2 and T3.

We then explore the difference between evaluating ATE-BERT on entire space and T1. ATE-BERT is trained on T1. Table~\ref{ATE-evaluation} presents the experimental results. We can see that the performance of ATE-BERT on T1 is much higher than that on entire space. Thus, evaluating ATE-BERT on T1 overestimates the performance of ATE-BERT.

\begin{figure}
	\centering
	\includegraphics[scale=0.35]{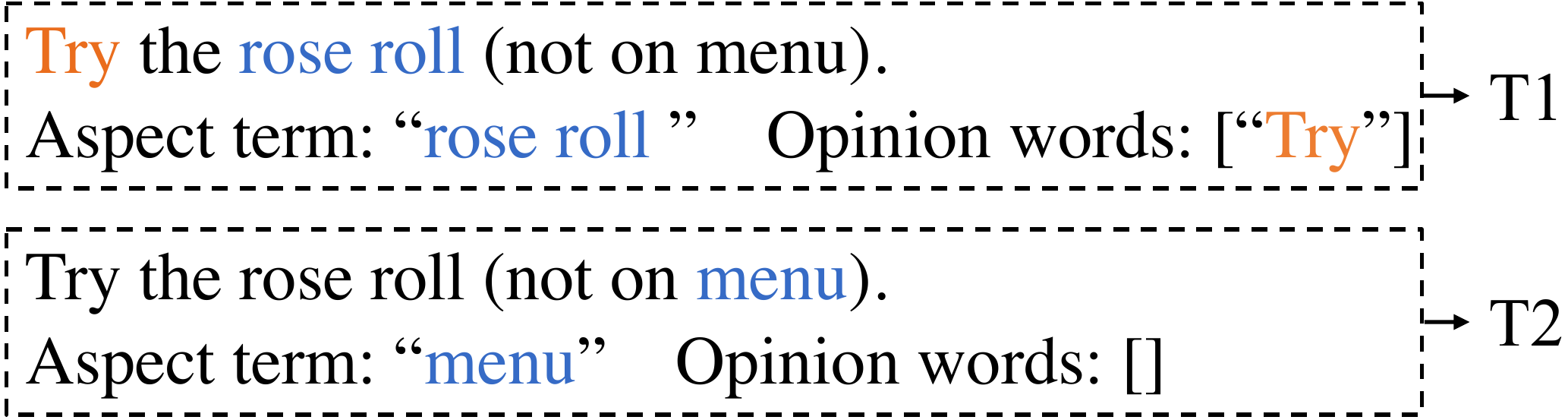}
	\caption{The two types of TOWE instances.}
	\label{fig:TOWE-instance-types}
\end{figure}

\subsubsection{TOWE Task}
The sample selection bias problem of Target-oriented Opinion Words Extraction (TOWE) models trained on non-entire-space datasets has been studied by \citet{10.1145/3477495.3531768}, therefore, we only give a brief description about their study and more details can be found in their paper.

As shown in Figure~\ref{fig:TOWE-instance-types}, there are two types of TOWE instances: T1 where aspect terms has corresponding opinion terms and T2 where aspect terms are not associated with any opinion terms. Non-entire-space datasets only include T1 instances. Entire space includes both T1 and T2.  \citet{10.1145/3477495.3531768} showed that evaluating TOWE models on non-entire-space datasets overestimates the performance of TOWE models and TOWE models trained on entire space obtain better performance than them trained on non-entire-space datasets.

\section{Conclusion}
In this paper, we find that there exist five versions of Aspect Sentiment Triplet Extraction (ASTE) datasets and researchers are not in agreement about which version of datasets should be used for ASTE. We analyze the relation between the five versions of datasets and suggest using the entire-space version. In addition, experimental results demonstrate that evaluating models on non-entire-space datasets inflates model performance and models trained on entire-space datasets have better generalization performance. 

\section{Limitations}
In this paper, in order to make ASTE-related works easy to follow, we suggest all researchers use the entire-space version of ASTE datasets. However, we only run six representative ASTE models on the entire-space version of datasets and report their performance. There are many other ASTE models. To completely achieve our goal and better help the research community, the experiments of other ASTE models on entire space should be carried out.

\bibliography{custom}
\bibliographystyle{acl_natbib}

\end{document}